\documentclass{svproc}
\usepackage{url}
\usepackage{graphicx}
\usepackage{longtable}
\usepackage{mathtools}
\usepackage{rotating}
\usepackage{tabularx}
\usepackage{ragged2e}
\usepackage{bm}
\usepackage{esvect}
\usepackage{kbordermatrix}
\usepackage{array}
\newcolumntype{P}[1]{>{\centering\arraybackslash}p{#1}}

\begin{document}
\mainmatter              
\title{The Power of Communities: A Text Classification Model with Automated Labeling Process Using Network Community Detection}
\titlerunning{The Power of Communities}  
%
\author{Minjun Kim\inst{1,2,4} \and Hiroki Sayama\inst{1,2,3}}
\authorrunning{Kim and Sayama.} 
%
\tocauthor{Minjun Kim, Hiroki Sayama}
\institute{Department of Systems Science and Industrial Engineering,\\
\and
Center for Collective Dynamics of Complex Systems,\\
Binghamton University, State University of New York, Binghamton, NY 13902, USA,\\
\and
Waseda Innovation Lab, Waseda University, Tokyo, Japan\\
\and
Pypestream, Inc., New York, USA\\
\email{mkim151@binghamton.edu}, \email{sayama@binghamton.edu}}

\maketitle              

\begin{abstract}
Text classification is one of the most critical areas in machine learning and artificial intelligence research. It has been actively adopted in many business applications such as conversational intelligence systems, news articles categorizations, sentiment analysis \cite{SEN09}, emotion detection systems \cite{EMO08}, and many other recommendation systems in our daily life. One of the problems in supervised text classification models is that the models' performance depends heavily on the quality of data labeling that is typically done by humans. In this study, we propose a new network community detection-based approach to automatically label and classify text data into multiclass value spaces. Specifically, we build networks with sentences as the network nodes and pairwise cosine similarities between the Term Frequency-Inversed Document Frequency (TFIDF) vector representations of the sentences as the network link weights. We use the Louvain method \cite{LOU08} to detect the communities in the sentence networks. We train and test the Support Vector Machine and the Random Forest models on both the human-labeled data and network community detection labeled data. Results showed that models with the data labeled by the network community detection outperformed the models with the human-labeled data by 2.68-3.75\% of classification accuracy. Our method may help developments of more accurate conversational intelligence and other text classification systems.

\end{abstract}
\section{Introduction}
Text data is a great source of knowledge for building many useful recommendation systems, search engines as well as conversational intelligence systems. However, it is often found to be a difficult and time-consuming task to structure the unstructured text data especially when it comes to labeling the text data for training text classification models. Data labeling, typically done by humans, is prone to make mislabeled data entries, and hard to track whether the data is correctly labeled or not. This human labeling practice indeed impacts on the quality of the trained models in solving classification problems. 

Some previous studies attempted to solve this problem by utilizing unsupervised \cite{YOU00,ZHO14,AND15} and semisupervised \cite{DOR16} machine learning models. However, those studies used a pre-defined keyword list for each category in the document, which provides the models with extra referential materials to look at when making the classification predictions, or included already labeled data as a part of the entire data set from which the models learn. In case of using unsupervised algorithms such as K$-$means and LDA \cite{ZHO14,AND15}, it is very much possible that frequently appearing words in multiple sentences can be used as features for multiple classes. This leads the models to render more ambiguity and to result in a poor performance in classifying documents. Also, the number of distinct classes ($K$) to be made is not determined systematically using the data, but heuristically by trying out many different values of $K$ which is not a reliable optimization.

Although there are many studies in text classification problems using machine learning techniques, there have been a limited number of studies conducted in text classifications utilizing network science. Network science is actively being adopted in studying biological networks, social networks, financial market prediction \cite{KSM17} and more in many fields of study to mine insights from the collectively inter-connected components by analyzing their relationships and structural characteristics. Only a few studies adopted network science theories to study text classifications and showed preliminary results of the text clustering performed by network analysis specially with the network community detection algorithms \cite{SAN08,MIK18,GER18}. However, those studies did not clearly show the quality of community detection algorithms or other possible useful features. Network community detection \cite{FOR16} is graph clustering methods used in complex networks analysis from large social networks analysis \cite{KAR08} to RNA-sequencing analysis \cite{KAN19} as a tool to partition a graph data into multiple parts based on the network's structural properties such as betweenness, modularity, etc.

In this paper, we study further to show the usefulness of the network community detection on labeling unlabeled text data that will automate and improve human labeling tasks, and on training machine learning classification models for a particular text classification problem. We finally show that the machine learning models trained on the data labeled by the network community detection model outperform the models trained on the human-labeled data.  

\section{Method}
\begin{figure}
	\centering
	\includegraphics[width=0.96\linewidth]{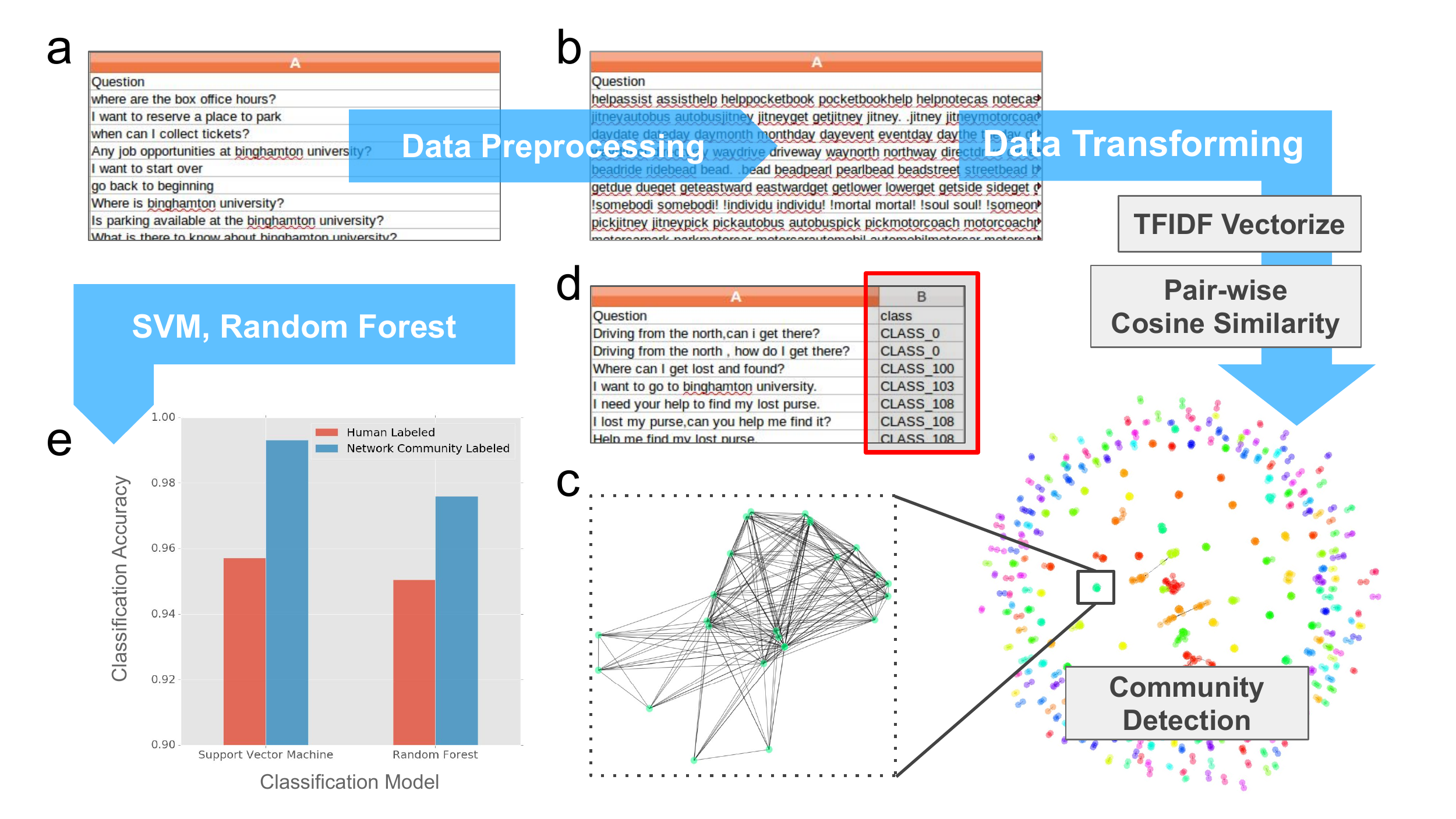}
	\caption{\textbf{Analysis process.} \textbf{a.} preprocess the text data by removing punctuations, stopwords and special characters, and add synonyms and bigrams, \textbf{b.} transform the prepocessed sentence into TFIDF vector, and compute pair-wise cosine similairy between every sentence pair, \textbf{c.} construct the sentence networks, and apply the Louvain method to detect communities of every sentence, \textbf{d.} label each sentence with the detected communities, \textbf{e.} train and test the Support Vector Machine and the Random Forest models on the labeled data.}
	\label{fig:diagram}
\end{figure}

We propose a new approach of building text classification models using a network community detection algorithm with unlabeled text data, and show that the network community detection is indeed useful in labeling text data by clustering the text data into multiple distinctive groups and also in improving the classification accuracy. This study takes the following steps (see Figure \ref{fig:diagram}), and uses Python packages such as {\fontfamily{qcr}\selectfont NLTK}, {\fontfamily{qcr}\selectfont NetworkX} and {\fontfamily{qcr}\selectfont SKlearn}.

\begin{itemize}
	\item Gathered a set of text data that was used to develop a particular conversational intelligence system from an artificial intelligence company, Pypestream. The data contains over 2,000 sentences of user expressions on that particular chatbot service such as 
	[``is there any parking space?", ``what movies are playing?", ``how can I get there if I'm taking a subway?"]\\
	
	\item Tokenizing and cleaning the sentences by removing punctuations, special characters and English stopwords that appear frequently without holding much important meaning. For example, [``how can I get there if I'm taking a subway?"] becomes [`get', `taking', `subway']\\
	
	\item Stemmizing the words following a suffix stripping algorithm \cite{POR06}, and adding synonyms and bigrams of the sequence of the words left in each sentence to enable the model to learn more kinds of similar expressions and the sequences of the words. For example, [`get', `taking', `subway'] becomes [`get', `take', `subway', `tube', `underground', `metro', `take metro', `get take', `take subway', `take underground', \dots]\\
	
	\item Transforming the preprocessed text data into a vector form by computing TFIDF of each preprocessed sentence with regard to the entire data set, and computing pair-wise cosine similiarity of the TFIDF vectors to form the adjacency matrix of the sentence network\\
	
	\item Constructing the sentence network using the adjacency matrix with each preprocessed sentence as a network node and the cosine similarity of TFIDF representations between every node pair as the link weight.\\
	
	\item Applying a network community detection algorithm on the sentence network to detect the communities where each preprocessed sentence belong, and build a labeled data set with detected communities for training and testing machine learning classification models.   
\end{itemize}

\subsection{Data, Preprocessing and Representation}
The data set obtained from Pypestream is permitted to be used for the research purpose only, and for a security reason, we are not allowed to share the data set. It was once originally used for creating a conversational intelligence system (chatbot) to support customer inquiries about a particular service. The data set is a two-column comma-separated value format data with one column of ``sentence" and the other column of ``class". It contains 2,212 unique sentences of user expressions asking questions and answering to the questions the chatbot asked the users. The sentences are all in English without having any misspelled words, and labeled with 19 distinct classes that are identified and designed by humans. Additional data set that only contains the sentences was made for this study by taking out the ``class" column from the original data set.

From each sentence, we removed punctuations, special characters and English stopwords to keep only those meaningful words that serve the main purpose of the sentence and to avoid any redundant computing. We then tokenized each sentence into words to process the data further in word level. For words in each sentence, we added synonyms of the words to handle more variations of the sentence as a typical method of increasing the resulting classification models' capability of understanding more unseen expressions with different words that describe similar meanings. The synonyms we added to the data are not context-specific synonyms, but all predefined synonyms of particular words. Although we used the predefined synonyms from the Python NLTK package, one might develop it's own synonym data to use under the context of the particular data to achieve better accuracy. We also added bigrams of the words to deal with those cases where the tokenization breaks the meaning of the word that consists of two words. For example, if we tokenized the sentence ``go to Binghamton University" and process the further steps without adding bigrams of them, the model is likely to yield lower confidence on classifying unseen sentences with ``Binghamton University" since the meaning of ``Binghamton University" is lost in the data set \cite{BEK04}.

With the preprocessed text data, we built vector representations of the sentences by performing weighted document representation using the TFIDF weighting scheme \cite{KAR12,TRS14,HUA08}. The TFIDF, as known as Term frequency inversed document frequency, is a document representation that takes account of the importance of each word by its frequency in the whole set of documents and its frequency in particular sets of documents. Specifically, let $D = \{d_1, \dots, d_n\}$ be a set of documents and $T = \{t_1, \dots, t_m\}$ the set of unique terms in the entire documents where $n$ is the number of documents in the data set and $m$ the number of unique words in the documents. In this study, the documents are the preprocessed sentences and the terms are the unique words in the preprocessed sentences. The importance of a word is captured with its frequency as $tf(d,t)$ denoting the frequency of the word $t \in T$ in the document $d \in D$. Then a document $d$ is represented as an $m$-dimensional vector $\vv{{t_d}}=(tf(d,t_1),\dots,tf(d,t_m))$. However, In order to compute more concise and meaningful importance of a word, TFIDF not only takes the frequency of a particular word in a particular document into account, but also considers the number of documents that the word appears in the entire data set. The underlying thought of this is that a word appeared frequently in some groups of documents but rarely in the other documents is more important and relevant to the groups of documents. Applying this concept, $tf(d,t)$ is weighted by the document frequency of a word, and $tf(d,t)$ becomes $tfidf(d,t) = tf(d,t)\times log\frac{|D|}{df(t)}$ where $df(t)$ is the number of documents the word $t$ appears, and thus the document $d$ is represented as $\vv{{t_d}}=(tfidf(d,t_1),\dots,tfidf(d,t_m))$.

\subsection{Sentence Network Construction}
\begin{figure}
	\centering
	\includegraphics[width=1.0\linewidth]{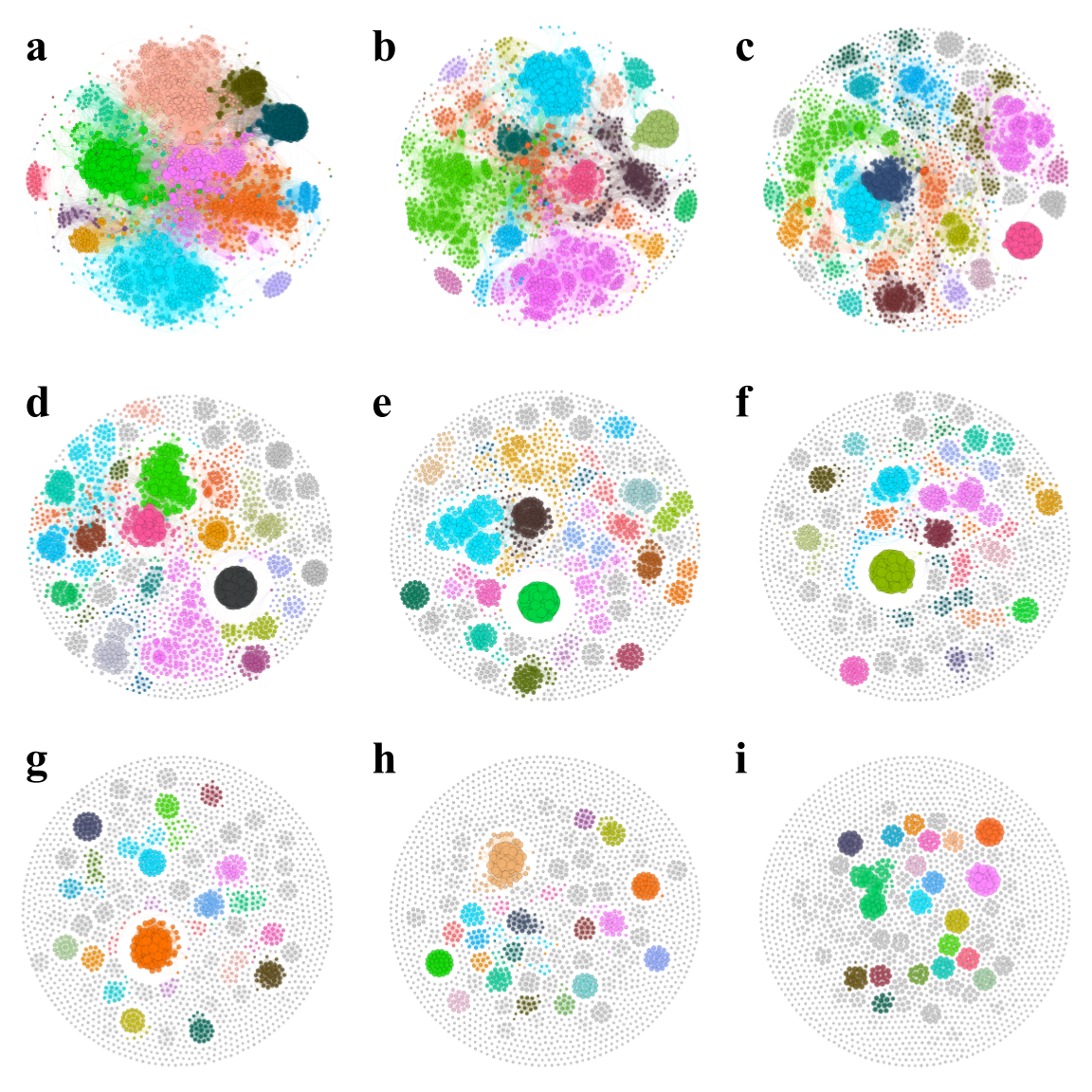}
	\caption{\textbf{Nine sentence networks with different connectivity thresholds.} Each node represents a sentence and an edge weight between two nodes represents the similarity between two sentences. In this study, we removed edges whose weight is below the threshold. \textbf{a.} network with threshold of 0.1 has 29 distinct communities with 11 signle node communities. \textbf{b.} network with threshold of 0.2 has 45 distinct communities with 20 single node communities, \textbf{c.} network with threshold of 0.3 has 100 distinct communities with 58 single node communities, \textbf{d.}network with threshold 0.4 has 187 distinct communities with 120 single node communities, \textbf{e.} network with threshold 0.5 has 320 distinct communities with 204 single node communities, \textbf{f.} network with threshold of 0.6 has 500 distinct communities with 335 single node communities, \textbf{g.} network with threshold of 0.7 has 719 distinct communities with 499 single node communities, \textbf{h.} network with threshold of 0.8 has 915 distinct communities with 658 single node communities, \textbf{i.} network with threshold of 0.9 has 1,140 distinct communities with 839 single node communities. Based on the visualized sentence networks, as the threshold gets larger it is shown that each network has more distinct communities.}
	\label{fig:community}
\end{figure}

With the TFIDF vector representations, we formed sentence networks. In total, 10 sentence networks (see Figure \ref{fig:threshold_00} and Figure \ref{fig:community}) were constructed with 2,212 nodes representing sentences and edge weights representing the pairwise similarities between sentences with 10 different network connectivity threshold values. The networks we formed were all undirected and weighted graphs. Particularly, as for the network edge weights, the cosine similarity \cite{HUA08,LIB13} is used to compute the similarities between sentences. The cosine similarity is a similarity measure that is in a floating number between 0 and 1, and computed as the angle difference between two vectors. A cosine similarity of $0$ means that the two vectors are perpendicular to each other implying no similarity, on the other hand, a cosine similarity of $1$ means that the two vectors are identical. It is popularly used in text mining and information retrieval techniques. In our study, the cosine similarity between two sentences $i$ and $j$ is defined as below equation.

\begin{equation}
SIM_C(\vv{t_{d_i}}, \vv{t_{d_j}}) = \frac{\vv{t_{d_i}} \cdot \vv{t_{d_j}}}{|\vv{t_{d_i}}|  |\vv{t_{d_j}}|}
\end{equation}
where:

$\vv{t_{d_i}} = (tfidf(d_i,t_1),\dots,tfidf(d_i,t_m))$, $the$ $TFIDF$ $vector$ $of$ $i$-$th$ $sentence$

$\vv{t_{d_j}} = (tfidf(d_j,t_1),\dots,tfidf(d_j,t_m))$, $the$ $TFIDF$ $vector$ $of$ $j$-$th$ $sentence$

$d$ $=$ $a$ $preprocessed$ $sentence$ $in$ $the$ $data$ $set$

$t$ $=$ $a$ $unique$ $word$ $appeared$ $in$ $the$ $preprocessed$ $data$ $set$\\

To build our sentence networks, we formed a network adjacency matrix for 2,212 sentences, $M$, with the pairwise cosine similarities of TFIDF vector representations computed in the above step. 

\renewcommand{\kbldelim}{(}
\renewcommand{\kbrdelim}{)}
\[
 M = \kbordermatrix{
	& d_1 & d_2 & \dots & d_{2212} \\
	d_1 & SIM_C(\vv{t_{d_1}},\vv{t_{d_1}}) & SIM_C(\vv{t_{d_1}},\vv{t_{d_2}}) & \dots &  SIM_C(\vv{t_{d_1}},\vv{t_{d_{2212}}})\\
	d_2 & SIM_C(\vv{t_{d_2}},\vv{t_{d_1}}) & SIM_C(\vv{t_{d_2}},\vv{t_{d_2}}) & \dots & SIM_C(\vv{t_{d_2}},\vv{t_{d_{2212}}}) \\
	\vdots & \vdots & \vdots & \ddots & \vdots \\
	d_{2212} & SIM_C(\vv{t_{d_{2212}}},\vv{t_{d_1}}) & SIM_C(\vv{t_{d_{2212}}},\vv{t_{d_2}}) & \dots & SIM_C(\vv{t_{d_{2212}}},\vv{t_{d_{2212}}})
}
\]

\subsection{Network Community Detection and Classification Models}
\begin{figure}
	\centering
	\includegraphics[width=0.6\linewidth]{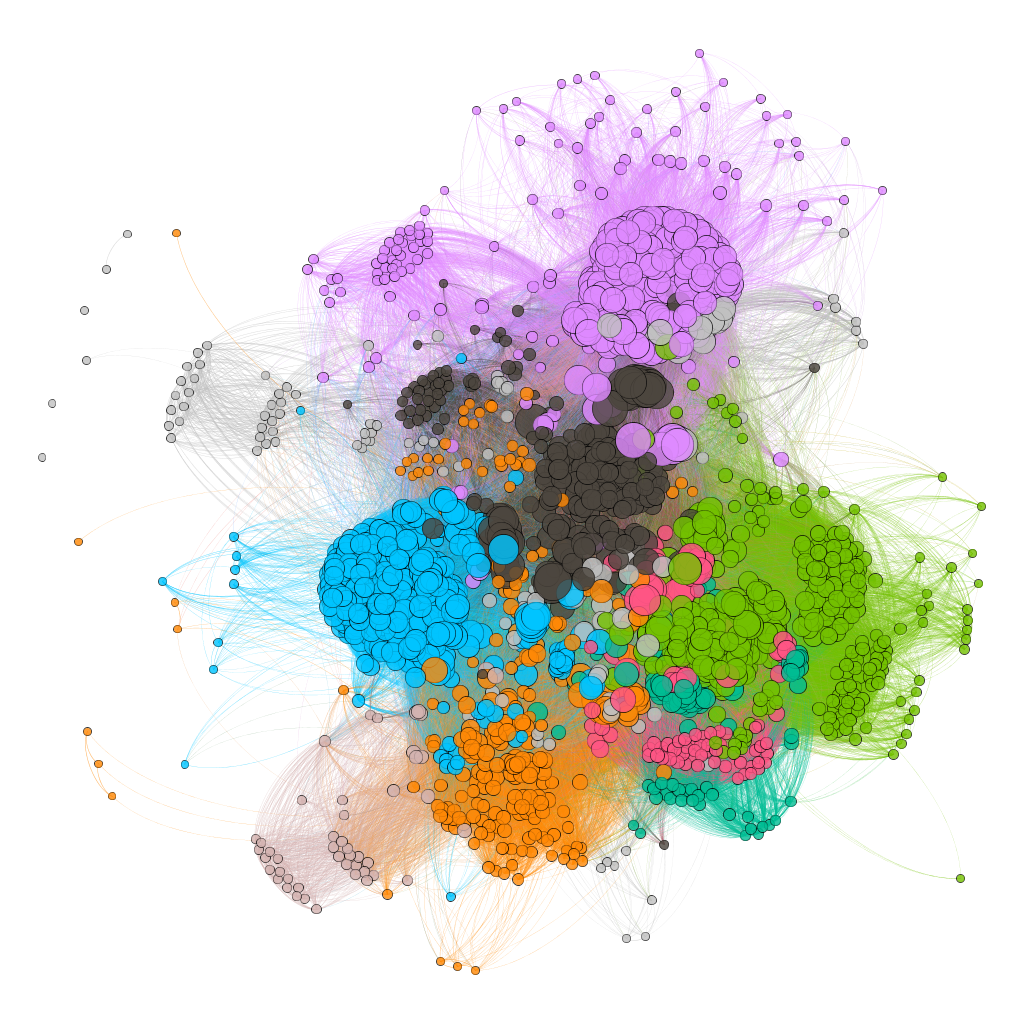}
	\caption{\textbf{A sentence network and its communities.} The sentence network with no threshold on the node connectivity has 18 distinct communities including three single node communities.}
	\label{fig:threshold_00}
\end{figure}

The particular algorithm of network community detection used in this study is the Louvain method \cite{LOU08} which partitions a network into the number of nodes - every node is its own community, and from there, clusters the nodes in a way to maximize each cluster's modularity which indicates how strong is the connectivity between the nodes in the community. This means that, based on the cosine similarity scores - the networks edge weights, the algorithm clusters similar sentences together in the same community while the algorithm proceeds maximizing the connectivity strength amongst the nodes in each community. The network constructed with no threshold in place was detected to have 18 distinct communities with three single node communities. Based on the visualized network (see Figure \ref{fig:threshold_00}), it seemed that the network community detection method clustered the sentence network as good as the original data set with human-labeled classes although the communities do not look quite distinct. However, based on the fact that it had three single node communities and the number of distinct communities is less than the number of classes in the human-labeled data set, we suspected possible problems that would degrade the quality of the community detection for the use of training text classification models.

\subsubsection{Quality of Network Community Detection Based Labeling} 
\begin{figure}
	\centering
	\includegraphics[width=0.8\linewidth]{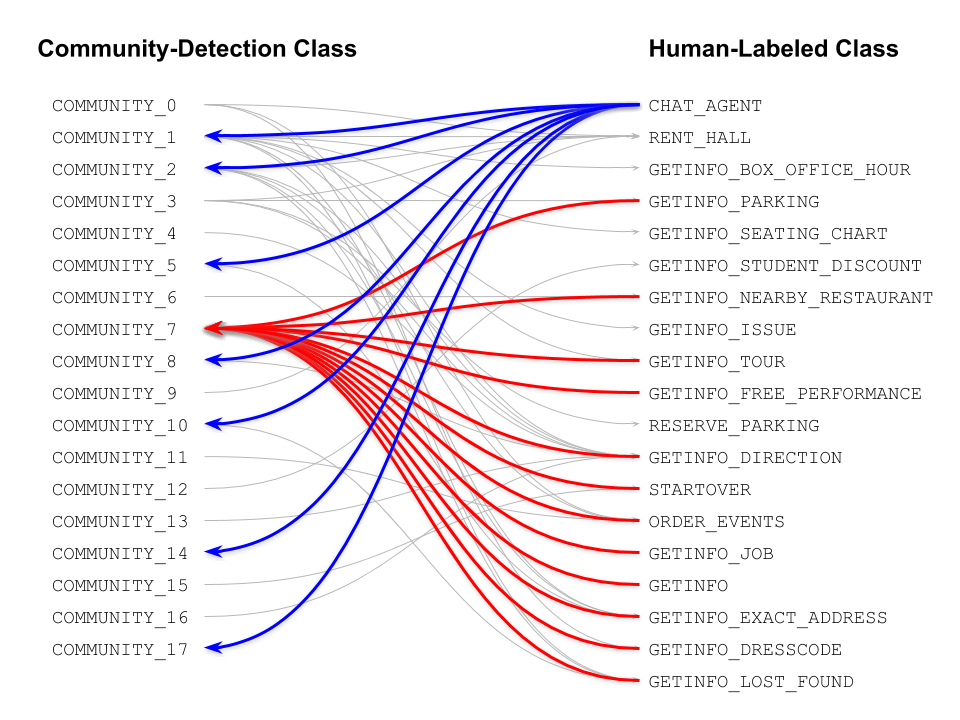}
	\caption{\textbf{A class map between detected communities and human-labeled classes.} The class map shows a mapping (all lines) between communities detected by the Louvain method and their corresponding human-labeled classes of the sentence network with no threshold.}
	\label{fig:community_labels}
\end{figure}

We checked the community detection results with the original human-labeled data by comparing the sentences in each community with the sentences in each human-labeled class to confirm how well the algorithm worked. We built class maps to facilitate this process (see Figure \ref{fig:community_labels}) that show mapping between communities in the sentence networks and classes in the original data set. Using the class maps, we found two notable cases where; 1. the sentences from multiple communities are consist of the sentences of one class of the human-labeled data, meaning the original class is splitted into multiple communities and 2. the sentences from one community consist of the sentences of multiple classes in human-labeled data, meaning multiple classes in the original data are merged into one community. For example, in the earlier case (see blue lines in Figure \ref{fig:community_labels}) which we call Class-split, the sentences in {\fontfamily{qcr}\selectfont COMMUNITY\_1}, {\fontfamily{qcr}\selectfont COMMUNITY\_2}, {\fontfamily{qcr}\selectfont COMMUNITY\_5}, {\fontfamily{qcr}\selectfont COMMUNITY\_8}, {\fontfamily{qcr}\selectfont COMMUNITY\_10}, {\fontfamily{qcr}\selectfont COMMUNITY\_14} and {\fontfamily{qcr}\selectfont COMMUNITY\_17} are the same as the sentences in {\fontfamily{qcr}\selectfont CHAT\_AGENT} class. Also, in the later case (see red lines in Figure \ref{fig:community_labels}) which we call Class-merge, the sentences in {\fontfamily{qcr}\selectfont COMMUNITY\_7} are the same as the sentences in {\fontfamily{qcr}\selectfont GETINFO\_PARKING}, {\fontfamily{qcr}\selectfont GETINFO\_NEARBY\_RESTAURANT}, {\fontfamily{qcr}\selectfont GETINFO\_TOUR}, {\fontfamily{qcr}\selectfont GETINFO\_EXACT\_ADDRESS}, {\fontfamily{qcr}\selectfont STARTOVER}, {\fontfamily{qcr}\selectfont ORDER\_EVENTS}, {\fontfamily{qcr}\selectfont GETINFO\_JOB}, {\fontfamily{qcr}\selectfont GETINFO}, {\fontfamily{qcr}\selectfont GETINFO\_DRESSCODE}, {\fontfamily{qcr}\selectfont GETINFO\_LOST\_FOUND} as well as {\fontfamily{qcr}\selectfont GETINFO\_FREE\_PERFORMANCE}.

The Class-split happens when a human-labeled class is divided into multiple communities as the sentence network is clustered based on the semantic similarity. This actually can help improve the text classification based systems to perform more sophisticatedly as the data set has more detailed subclasses to structure the systems with. Although it is indeed a helpful phenomenon, we would like to minimize the number of subclasses created by the community detection algorithm simply because we want to avoid having too many subclasses that would add more complexity in designing any applications using the community data. On the other hand, the Class-merge happens when multiple human-labeled classes are merged into one giant community. This Class-merge phenomenon also helps improve the original data set by detecting either mislabeled or ambiguous data entries. We will discuss more details in the following subsection. Nonetheless, we also want to minimize the number of classes merged into the one giant community, because when too many classes are merged into one class, it simply implies that the sentence network is not correctly clustered. For example, as shown in Figure \ref{fig:community_labels} red lines, 12 different human-labeled classes that do not share any similar intents are merged into {\fontfamily{qcr}\selectfont COMMUNITY\_7}. If we trained a text classification model on this data, we would have lost the specifically designed purposes of the 12 different classes, expecting {\fontfamily{qcr}\selectfont COMMUNITY\_7} to deal with all the 12 different types of sentences. This would dramatically degrade the performance of the text classification models.

\begin{figure}
	\centering
	\includegraphics[width=0.7\linewidth]{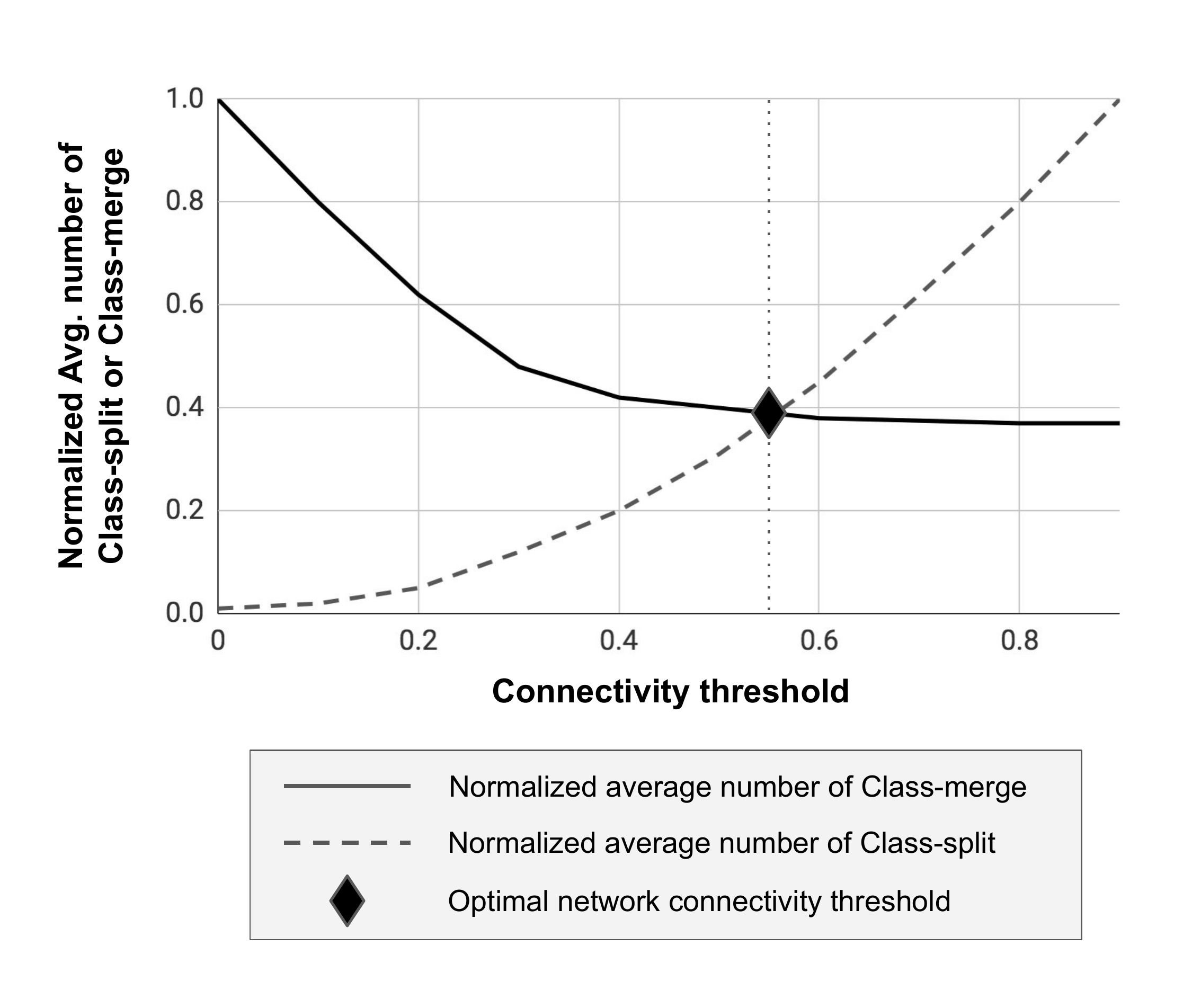}
	\caption{\textbf{Optimal connectivity threshold point based on Class-split and Class-merge mertics} The normalized Class-split score (blue line) increases as the threshold gets larger. On the other hand, normalized Class-merge (red line) decreases as the threshold gets larger. The optimal connectivity threshold is the point where both scores are minimized which is 0.5477.}
	\label{fig:class_map}
\end{figure}

In order to quantify the degree of Class-split and Class-merge of a network, and to find out optimal connectivity threshold that would yield the sentence network with the best community detection quality, we built two metrics using the class map. We quantified the Class-split by counting the number of communities splitted out from each and every human-labeled class, and the Class-merge by counting the number of human-labeled classes that are merged into each and every community. We then averaged the Class-splits across all the human-labeled classes and Class-merges across all the communities. For example, using the class map of the sentence network with no threshold, we can easily get the number of Class-split and Class-merge as below. By averaging them, we get the Class\_split and Class\_merge scores of the sentence network, which is 2.7368 and 2.8333 respectively.

\renewcommand{\kbldelim}{(}
\renewcommand{\kbrdelim}{)}
\[
Class\_split = [2,1,4,5,1,2,2,1,1,4,1,9,1,1,4,2,2,2,7]
\]
\renewcommand{\kbldelim}{(}
\renewcommand{\kbrdelim}{)}
\[
Class\_merge=[1,1,1,1,4,1,2,1,2,4,1,9,2,1,1,1,6,12]
\]

We computed the normalized Class\_split and Class\_merge scores for all 10 sentence networks (see Figure \ref{fig:class_map}). Figure \ref{fig:class_map} shows the normalized Class-split and Class-merge scores of the 10 sentence networks with different connectivity thresholds ranging from $0.0$ to $0.9$. With these series of Class\_split and Class\_merge scores, we found out that at 0.5477 of connectivity threshold we can get the sentence network that would give us the best quality of community detection result particularly for our purpose of training text classification models.

\subsubsection{Detecting Mislabeled or Ambiguous Sentences in Human-made Data Set} Using the Class\_merge information we got from the class map, we were able to spot out those sentences that are either mislabeled or ambiguous between classes in the original data set. This is extreamly helpful and convenient feature in fixing and improving text data for classification problems, because data fixing is normally a tedious and time consuming task which takes a great amount of human labor. For example, by looking at the class map, in our sentence network with no threshold, {\fontfamily{qcr}\selectfont COMMUNITY\_5} contains sentences appeared in {\fontfamily{qcr}\selectfont GETINFO\_EXACT\_ADDRESS} and {\fontfamily{qcr}\selectfont CHAT\_AGENT} classes. We investigated the sentences in {\fontfamily{qcr}\selectfont COMMUNITY\_5}, and were able to spot out one sentence [``I need to address a human being!"] which is very ambiguous for machines to classify between the two classes. This sentence is originally designed for {\fontfamily{qcr}\selectfont CHAT\_AGENT} class, but because of its ambiguous expression with the word `{\fontfamily{qcr}\selectfont address}', it is together with sentences in {\fontfamily{qcr}\selectfont GETINFO\_EXACT\_ADDRESS} in {\fontfamily{qcr}\selectfont COMMUNITY\_5}. After fixing the ambiguity of that sentence by correcting it to [``I need to talk to a human being!''], we easily improved the original data set.

\subsubsection{Classification Models}
Once we got the optimal connectivity threshold using the Class\_split and Class\_merge scores as shown in previous sections, we built the sentence network with the optimal threshold of 0.5477. We then applied the Louvain method to detect communities in the network, and to automatically label the data set. The network with the threshold of 0.5477 has 399 communities with 20,856 edges. Class\_split and Class\_merge scores of the network were 22.3158 and 1.0627 respectively. We finally trained and tested machine learning based text classification models on the data set labeled by the community detection outcome to see how well our approach worked. Following a general machine learning train and test practice, we split the data set into a train set (80\% of the data) and a test set (20\% of the data). The particular models we trained and tested were the Support Vector Machine \cite{DRU99} and the Random Forest \cite{WUQ14} models that are popularly used in natural language processing such as spam e-mail and news article categorizations. More details about the two famous machine learning models are well discussed in the cited papers.

\begin{table}
	\caption{\textbf{Accuracies of text classification models} It is shown that the models trained on the community data resulted in higher accuracy in classifying the sentences in the test data.}
	\begin{center}
		\bgroup
		\def\arraystretch{1.3}
		\begin{tabular} { | p{4.5cm} | P{1.5cm} | P{2.5cm} | } 
			\hline
			Data Labeling & SVM & Random forest \\[1pt]
			\hline
			Human Labeled & 0.9572 &  0.9504 \\ [1pt]
			Network Community Labeled & 0.9931 & 0.9759 \\ [1pt]
			\hline
		\end{tabular}
		\label{table:accuracy_plot}
		\egroup
	\end{center}
\end{table}


\section{Result}
Table \ref{table:accuracy_plot} shows the accuracy of the four Support Vector Machine and the Random Forest models trained on the original human-labeled data and the data labeled by our method. The accuracies are hit ratios that compute the number of correctly classified sentences over the number of all sentences in the test data. For example, if a model classified 85 sentences correctly out of 100 test sentences, then the accuracy is 0.85. In order to accurately compute the ground truth hit ratio, we used the ground truth messages in the chatbot. The messages are the sentences that are to be shown to the chatbot users in response to the classification for a particular user query as below.

\renewcommand{\kbldelim}{(}
\renewcommand{\kbrdelim}{)}
\[
input \hspace{0.1cm} sentence \rightarrow detected \hspace{0.1cm} class \rightarrow output \hspace{0.1cm} message
\]

For example, for a question of ``{\fontfamily{qcr}\selectfont how do I get there by subway?}", in the chatbot, there is a designed message of ``{\fontfamily{qcr}\selectfont You can take line M or B to 35th street}" to respond to that particular query. Using these output messages in the chatbot, we were able to compute the ground truth accuracy of our classification models by comprehending the input sentences in the test sets, the detected classes from the models and linked messages. In our test, the Support Vector Machine trained on human-labeled data performed 0.9572 while the same model trained on the data labeled by our method resulted in 0.9931. Also, the Random Forest model trained on human-labeled data resulted in an accuracy value of 0.9504 while the same model trained on the data labeled by our method did 0.9759.

\section{Discussions and Conclusion}

In this study, we demonstrated a new approach of training text classification models using the network community detection, and showed how the network community detection can help improve the models by automatically labeling text data and detecting mislabeled or ambiguous data points. As seen in this paper, we were able to yield better results in the accuracy of the Support Vector Machine and the Random Forest models compared to the same models that were trained on the original human-labeled data for the particular text classification problem. Our approach is not only useful in producing better classification models, but also in testing the quality of human-made text data. One might be able to get even better results using this method by utilizing more sophisticatedly custom-designed synonyms and stopwords, using more advanced natural language processing methods such as word-embeddings, utilizing higher n-grams such as trigrams, and using more balanced data sets. In the future, we would like to expand this study further to use the network itself to parse out classifications of unseen sentences without training machine learning models.

%
%

\end{document}